%% file: main.tex
\documentclass[11pt,a4paper]{article}
\usepackage[hyperref]{emnlp2018}
\usepackage{times}
\usepackage{latexsym}
\usepackage{makecell}
\usepackage{graphicx}
\usepackage{subfigure}
\usepackage{algorithm}
\usepackage{amssymb}
\usepackage{amsmath}
\usepackage[noend]{algpseudocode}
\usepackage{multirow}
\usepackage{array}
\usepackage{tabularx}
\usepackage{booktabs}

\aclfinalcopy

\usepackage{url}

\input{defs}

\newcommand{\namecite}[1]{\newcite{#1}}

\newcommand{\textst}[1]{\texttt{\small #1}}

\title{Multi-Reference Training with Pseudo-References \\ for Neural Translation and Text Generation}

\author{
  Renjie Zheng $^1$
  \qquad
  Mingbo Ma $^{1,2}$
  \qquad
  Liang Huang $^{1,2}$
\\ 
$^{1}$School of EECS, Oregon State University, Corvallis, OR \\
$^{2}$Baidu Research, Sunnyvale, CA\\
  {\small \tt zheng@renj.me \quad \{cosmmb, liang.huang.sh\}@gmail.com} \\
  }
\date{}

\begin{document}
\maketitle
\begin{abstract}
Neural text generation, including neural machine translation, image captioning, and summarization, has been quite successful
recently. However, during training time, 
typically only one reference is considered for each example,
even though there are often multiple references available,
e.g., 4 references in NIST MT evaluations, and 5 references in image captioning data.
We first investigate several different ways of utilizing multiple human references during training.
But more importantly, we then propose an algorithm to generate 
exponentially many pseudo-references 
by first compressing existing human references into lattices
and then traversing them to generate new pseudo-references.
These approaches lead to substantial improvements over strong baselines in both machine translation (+1.5 BLEU) and image captioning (+3.1 BLEU / +11.7 CIDEr).
\end{abstract}

\input{intro}

\input{method}

\input{lattice}

\input{exps}

\section{Conclusions}
We introduce several multiple-reference training methods and 
a neural-based lattice compression framework, which can generate more training references based on existing ones.
Our proposed framework outperforms the baseline models on both MT and image captioning tasks.

\section*{Acknowledgments}

This work was supported in part by DARPA grant N66001-17-2-4030, and NSF grants IIS-1817231 and IIS-1656051.
We thank the anonymous reviewers for suggestions and Juneki Hong for proofreading.

\bibliography{emnlp2018}
\bibliographystyle{acl_natbib_nourl}

\appendix

\end{document}

%% file: defs.tex
\newcommand{\notes}[1]{}

\newcommand{\vech}{\ensuremath{\mathbf{h}}}

\newcommand{\vecy}{\ensuremath{\mathbf{y}}}
\newcommand{\vecx}{\ensuremath{\mathbf{x}}}

\newcommand{\ith}[1]{\ensuremath{i^{{th}}}}

\newcount\permx
\newcount\permy
\def\permdot#1#2{
\permx=#1 \advance\permx by-1
\permy=#2 \advance\permy by-1
\psframe[fillcolor=black, fillstyle=solid]
(\permx,\permy)(#1, #2)
}

\newcommand{\boxnum}[1]{{\setlength{\fboxsep}{1pt}\raisebox{1pt}{\hspace{1pt}\fbox{\tiny #1}\hspace{1pt}}}}
\newcommand{\ind}[1]{\ensuremath{_{\kern-0.5pt\boxnum{#1}}}}

\newcommand{\smallnt}[1]{\ensuremath{_{\mbox{\tiny PP}}}\xspace}

\newcommand{\pseudocode}{Algorithm}
\floatname{algorithm}{\pseudocode}

\iffalse

\else

\fi



%% file: intro.tex
\section{Introduction}

Neural text generation has attracted much attention in recent years thanks to 
its impressive generation accuracy and wide applicability. In addition to demonstrating compelling 
results for machine translation (MT) \cite{Sutskever:2014, Bahdanau:14}, by simple adaptation, 
practically very same or similar models 
have also proven to be successful for 
summarization \cite{Rush:15,Nallapati:16} and image or video captioning \cite{Venu:15,Kelvin:15}.

The most common neural text generation model is based on the encoder-decoder framework \cite{Sutskever:2014} 
which generates a variable-length output sequence using an RNN-based decoder with 
attention mechanisms \cite{Bahdanau:14,Xu:2015}. 
There are many recent efforts in improving the generation accuracy, e.g., ConvS2S  \cite{Gehring:2017} and Transformer  \cite{Vaswani:2017}. 
However, all these efforts are limited to training with a single reference even when multiple references are available.

\if
While the RNN-based generation models 
require sequential order during training and decoding,
there are also alternative approaches that do not,
including
the ConvS2S model \cite{Gehring:2017} 
is purely based on CNNs, 
the Transformer model \namecite{Vaswani:2017} 
that only uses attention mechanisms 
without recurrence or convolution.

Though all these text generation 
models could achieve competitive accuracy, 
they still suffer from a crucial limitation:
during training, they only consider one single reference and discard the other references 
when there are multiple available.
\fi

Multiple references are essential for evaluation 
due to the non-uniqueness of translation and generation unlike classification tasks.
In MT, even though the training sets are usually with single reference (bitext), the evaluation sets often come with multiple references. 
For example, the NIST Chinese-to-English and Arabic-to-English MT evaluation datasets (2003--2008) have in total around 10,000 
Chinese sentences and 10,000 
Arabic sentences each with 4 different English translations. 
On the other hand, 
for image captioning datasets,
multiple references are more 
common not only for evaluation, but also for training, e.g., the MSCOCO \cite{MSCOCO} dataset provides 5 references per image
and PASCAL-50S and ABSTRACT-50S \cite{Vedantam:15} even provide 50 references per image. 
Can we use the extra references during training? How much can we benefit from training with multiple references? 


We therefore first investigate several different ways of
utilizing existing human-annotated references, which include Sample One~\cite{karpathy2015deep}, Uniform, and Shuffle methods
(explained in Sec.~\ref{sec:train}). 
Although Sample One has been explored in image captioning,
to the best of our knowledge, this is the first time that an MT system is trained with multiple references. 

Actually, four or five references still cover only a tiny fraction 
of the exponentially large space of potential references \cite{dreyer2012hyter}.
More importantly, encouraged by the success of training with multiple  human 
references, 
we further propose a framework to generate many more pseudo-references automatically. 
In particular, we design a neural multiple-sequence alignment algorithm to compress all existing human 
references into a lattice by merging similar words across different references (see examples in Fig.~\ref{fig:lattice});
this can be viewed as a modern, neural version of paraphrasing with multiple-sequence alignment \cite{Barzilay:2003,Barzilay:2002}.
We can then generate theoretically exponentially more references from the lattice. 


We make the following main contributions:

\begin{itemize}
\item Firstly, we investigate three different methods for multi-reference training on both MT and image captioning tasks (Section \ref{sec:train}).
\item Secondly, we propose a novel neural network-based multiple sequence alignment model to compress the existing references into lattices. By traversing these lattices, we generate exponentially many new pseudo-references (Section \ref{sec:lattice}).
\item We report substantial improvements over strong baselines in both MT (+1.5 BLEU) and image captioning (+3.1 BLEU / +11.7 CIDEr) by training on the newly generated pseudo-references (Section \ref{sec:experiments}).
\end{itemize}

%% file: method.tex

\section{Using Multiple References}
\label{sec:train}

In order to make the multiple reference training easy to adapt to any frameworks, we do not change anything from the existing models itself.
Our multiple reference training is achieved by converting a multiple reference dataset to a single reference dataset without losing any information.

Considering a multiple reference dataset $D$,
where the $i^{th}$ training example, $( \vecx_i , Y_i )$,
includes one source input $\vecx_i$, which is a source sentence
in MT or image vector in image captioning,
and a reference set $Y_i = \{\vecy_i^1, \vecy_i^2, ... \vecy_i^K\}$
of $K$ references. We have the following methods to convert the multiple reference dataset to a single reference dataset $D'$
(note that the following $D_{\text{sample one}}'$, $D_{\text{uniform}}'$
and $D_{\text{shuffle}}'$ are ordered sets):

\smallskip
\noindent{\textbf{Sample One}}:
The most straightforward way is to use a different reference in different epochs during training to explore the variances between references.
For each example, we randomly pick one of the $K$ references
in each training epoch (note that the random function will be used in each epoch).
This method is commonly used in existing image captioning literatures,
such as \cite{karpathy2015deep}, but never used in MT.
This approach can be formalized as:
$$D'_{\text{sample\ one}} = \bigcup_{i = 1}^{|D|} \{ (\vecx_i, \vecy_i^{k_{i}}) \}, k_i = \text{rand}(1, ..., K)$$



\noindent{\textbf{Uniform}}: Although all references are accessible
by using Sample One,
it is not guaranteed that all references are used during training.
So we introduce \textbf{Uniform} which basically
copies $\vecx_i$ training example $K$ times and each time with a different reference.
This approach can be formalized as:

$$D'_{\text{uniform}} = \bigcup_{i = 1}^{|D|} \bigcup_{k=1}^{K} \{ (\vecx_i, \vecy_i^{k}) \}$$

\noindent{\textbf{Shuffle}} is based on Uniform,
but shuffles all the source and reference pairs in random order before each epoch.
So, formally it is:

$$D'_{\text{shuffle}} = \text{Shuffle}(D'_{\text{uniform}})$$

Sample One is supervised by different training signals in different epochs while both Uniform and Shuffle include all the references at one time. 
Note that we use mini-batch during training. When we set the batch size equal to the entire training set size in both Uniform and Shuffle, they become equivalent.

%% file: lattice.tex

\section{Pseudo-References Generation}
\label{sec:lattice}

\begin{figure*}[!ht]
\centering
\subfigure{
\includegraphics[width=1.0\textwidth]{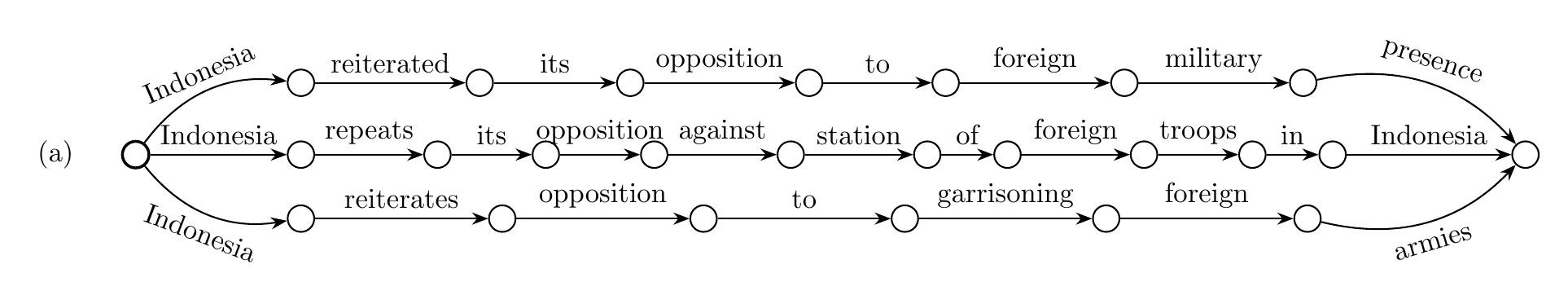}
\label{fig:lattice_init}
}
\hrule
\subfigure{
\includegraphics[width=1.0\textwidth]{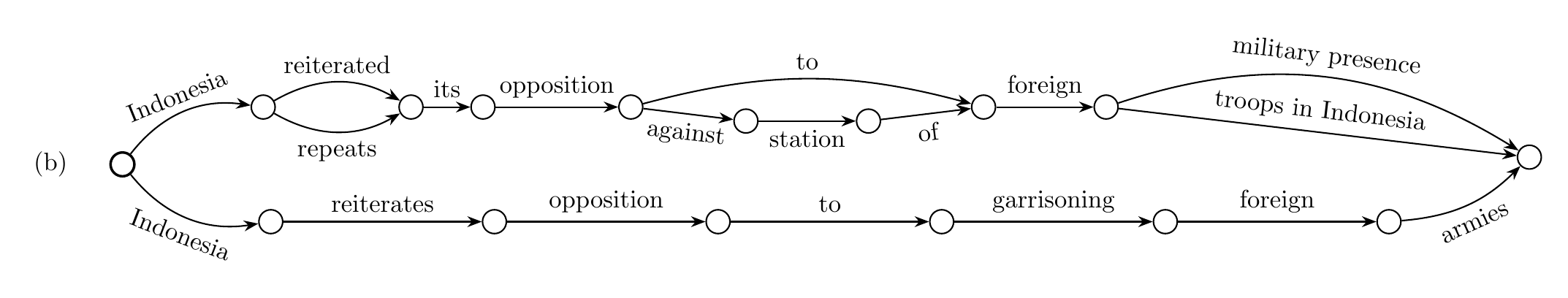}
\label{fig:lattice_2_hard}
}\\[-0.35cm]
\subfigure{
\includegraphics[width=1.0\textwidth]{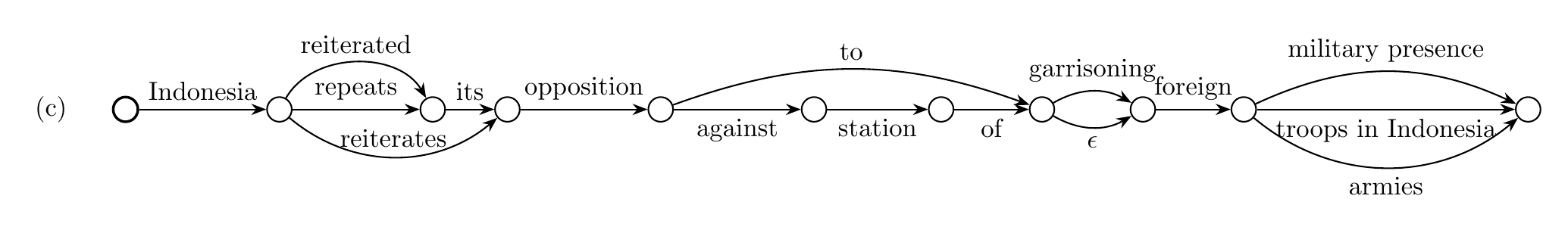}
\label{fig:lattice_3_hard}
}
\hrule
\subfigure{
\includegraphics[width=1.0\textwidth]{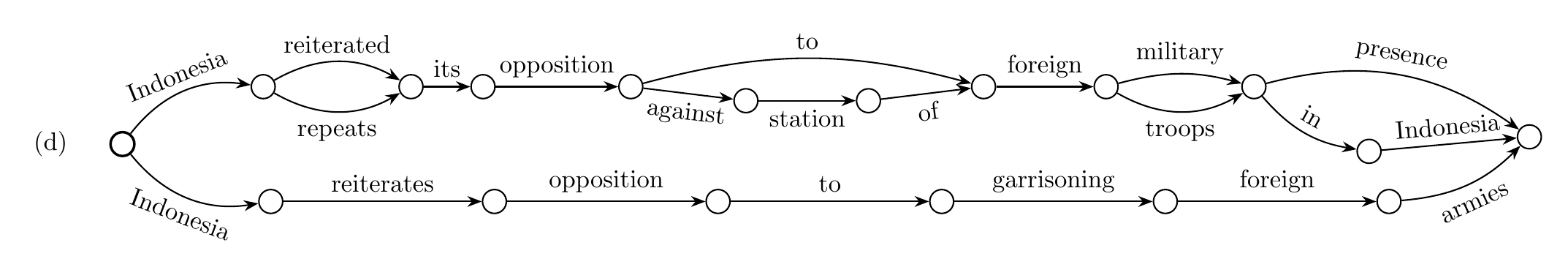}
\label{fig:latticeb}
}\\[-0.3cm]
\subfigure{
\includegraphics[width=1.0\textwidth]{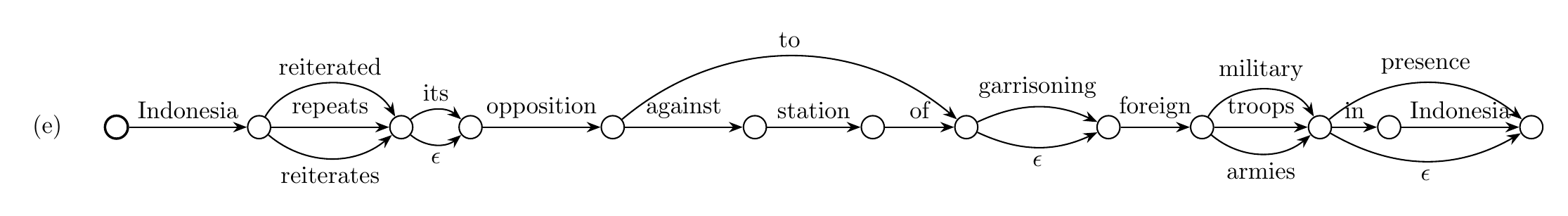}
\label{fig:latticec}
\vspace{-0.5cm}
}
\vspace{-3ex}
\caption{Lattice construction with word alignment. (b-c) is hard word alignment and 33 pseudo-references can be generated. (d-e) is soft word alignment, 213 pseudo-references can be generated.}

\label{fig:lattice}
\end{figure*}


In text generation tasks, the given multiple references are
only a small portion in the whole space of potential references.
To cover a larger number of references during training, we want to generate
more pseudo-references which is similar to existing ones.


Our basic idea is to compress different references $\vecy_0, \vecy_1, ..., \vecy_K$ into a lattice.
We achieve this by merging similar words in the references.
Finally, we generate more pseudo-references by simply traversing
the compressed lattice and select those with high quality according to its
BLEU score.

Take the following three references from the NIST 
Chinese-to-English machine translation dataset as an example:

\begin{enumerate}
	{\small
	\item \texttt{Indonesia reiterated its opposition to foreign military presence}
	\item \texttt{Indonesia repeats its opposition against station of foreign troops in Indonesia}
	\item \texttt{Indonesia reiterates opposition to garrisoning foreign armies}
	}
\end{enumerate}

\subsection {Naive Idea: Hard Word Alignment}

The simplest way to compress different references into a lattice
is to do pairwise reference compression iteratively.
At each time, we select two references and merge the same words
in them.

Considering the previous example, we can derive an initial
lattice from the three references as shown in Fig.~\ref{fig:lattice_init}.
Assume that we first do a pairwise reference compression on first two references,
we can merge at four sharing words: \textst{Indonesia}, \textst{its}, \textst{opposition} and \textst{foreign}, and the lattice will turn to
Fig.~\ref{fig:lattice_2_hard}. 
If we further compress the first and third references, we can merge at
\textst{Indonesia}, \textst{opposition}, \textst{to} and \textst{foreign},
which gives the lattice Fig.~\ref{fig:lattice_3_hard}.
By simply traversing the final lattice, 33 new pseudo-references
can be generated. For example:

\begin{enumerate}
	{\small
	\item \texttt{Indonesia reiterated its opposition to garrisoning foreign armies}
	\item \texttt{Indonesia repeats its opposition to foreign military presence}
	\item \texttt{Indonesia reiterates opposition to foreign troops in Indonesia}
	\item \texttt{ ... }
	}
\end{enumerate}


\begin{figure*}[!htbp]
\begin{minipage}[c][][t]{.25\linewidth}
  \centering
  \includegraphics[width=1.0\linewidth]{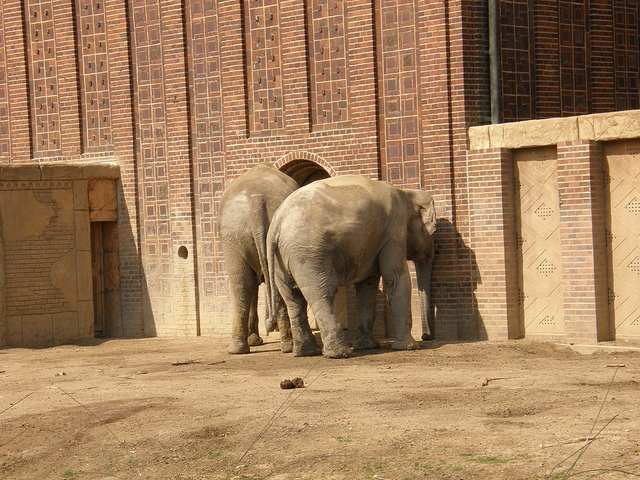}
  \label{fig:wrong_hard_image}
\end{minipage}%
\begin{minipage}[c][][t]{.75\linewidth}
  \centering
  \includegraphics[width=1.0\linewidth]{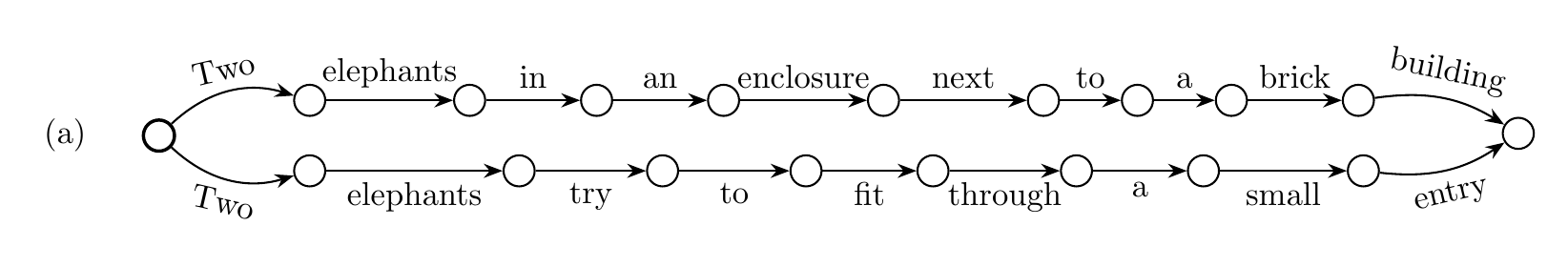}\\[-0.5cm]
  \label{fig:wrong_hard_lattice_initial}
  \includegraphics[width=1.0\linewidth]{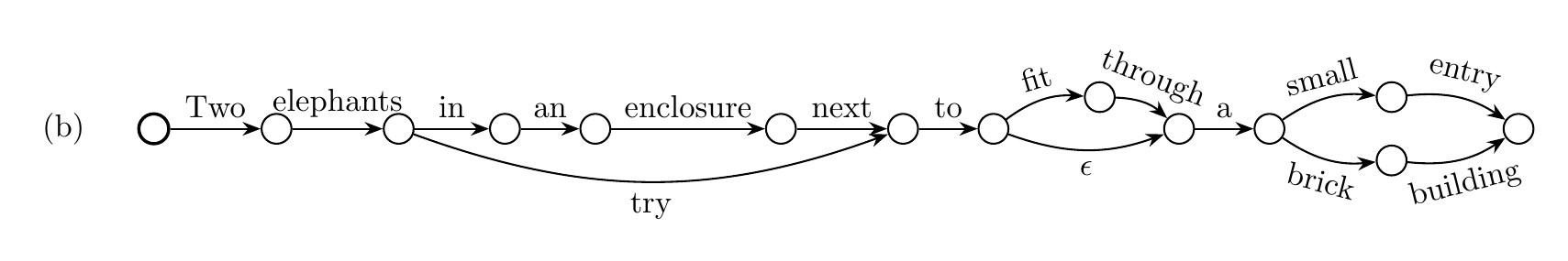}\\
  \label{fig:wrong_hard_lattice}
\end{minipage}
\vspace{-0.5cm}
\caption{Mistakes from hard word alignment by merging at ``\texttt{to}''.}
\label{fig:wrong_hard_alignment}
\end{figure*}

However, this simple hard alignment method
(only identical words can be aligned)
 suffers from two problems:

\begin{enumerate}
\item Different words may have similar meanings
and need to be merged together.
For example, in the previous example, \textst{reiterated},
\textst{repeats} and \textst{reiterates} should be merged together.
Similarly, \textst{military}, \textst{troops} and \textst{armies}
also have similar meanings.
If the lattice can align these words, we can generate the lattice
shown in Fig.~\ref{fig:latticec} which can generate 213 pseudo-references.

\item Identical words may have different meaning
in different contexts and should not be merged.
Considering the following two references from
the COCO image captioning dataset (corresponding picture is shown in Fig.~\ref{fig:wrong_hard_alignment}): 
\end{enumerate}

\begin{enumerate}
	{\small
	\item \texttt{Two elephants in an enclosure next to a brick building}
	\item \texttt{Two elephants try to fit through a small entry}
	}
\end{enumerate}


Following the previously described algorithm, we can merge
the two references at ``\textst{two elephants}'',
at ``\textst{to}'' and at ``\textst{a}''.
However, ``\textst{to}'' in the two references are very different
(it is a preposition in the first reference and an infinitive in the second)
and should not be merged.
Thus, the lattice in Fig.~\ref{fig:wrong_hard_alignment}(b) will generate
the following wrong pseudo-references:
\begin{enumerate}
	{\small
	\item \texttt{Two elephants try to a small entry}
	\item \texttt{Two elephants in an enclosure next to fit through a brick building}
	}
\end{enumerate}

Therefore, we need to investigate a better method to compress the lattice.

\input{alignment}

\input{dp}

\input{generate}

%% file: alignment.tex

\subsection{Measuring Word Similarity in Context}

To tackle the above listed two problems of hard alignment,
we need to identify synonyms and words with similar meanings.
\namecite{Barzilay:2002} utilize an external synonyms dictionary to get the similarity score between words.
However, this method ignores the given context of each word. 
For example, in Fig.~\ref{fig:lattice_init},
there are two \textst{Indonesia}'s in the second path of reference.
If we use a synonyms dictionary, both \textst{Indonesia} tokens will be aligned to the \textst{Indonesia} in the first or third sentence with the same score.
This incorrect alignment would lead to meaningless lattice.

Thus, we introduce the semantic substitution matrix which
measures the semantic similarity of each word pairs in context.
Formally, given a sentence pair $\vecy_i$ and $\vecy_j$,
we build a semantic substitution matrix
$M = \mathbb{R}^{|\vecy_i| \times |\vecy_j|}$,
whose cell $M_{u, v}$ represents the similarity score between word $\vecy_{i,u}$ and word $\vecy_{j,v}$.

We propose a new neural network-based multiple sequence alignment algorithm
to take context into consideration.
We first build a language model (LM) to obtain
the semantic representation of each word,
then these word representations
are used to construct the semantic substitution matrix between sentences.

\begin{figure}[b]
\begin{center}
\subfigure{
\includegraphics[width=8.0cm]{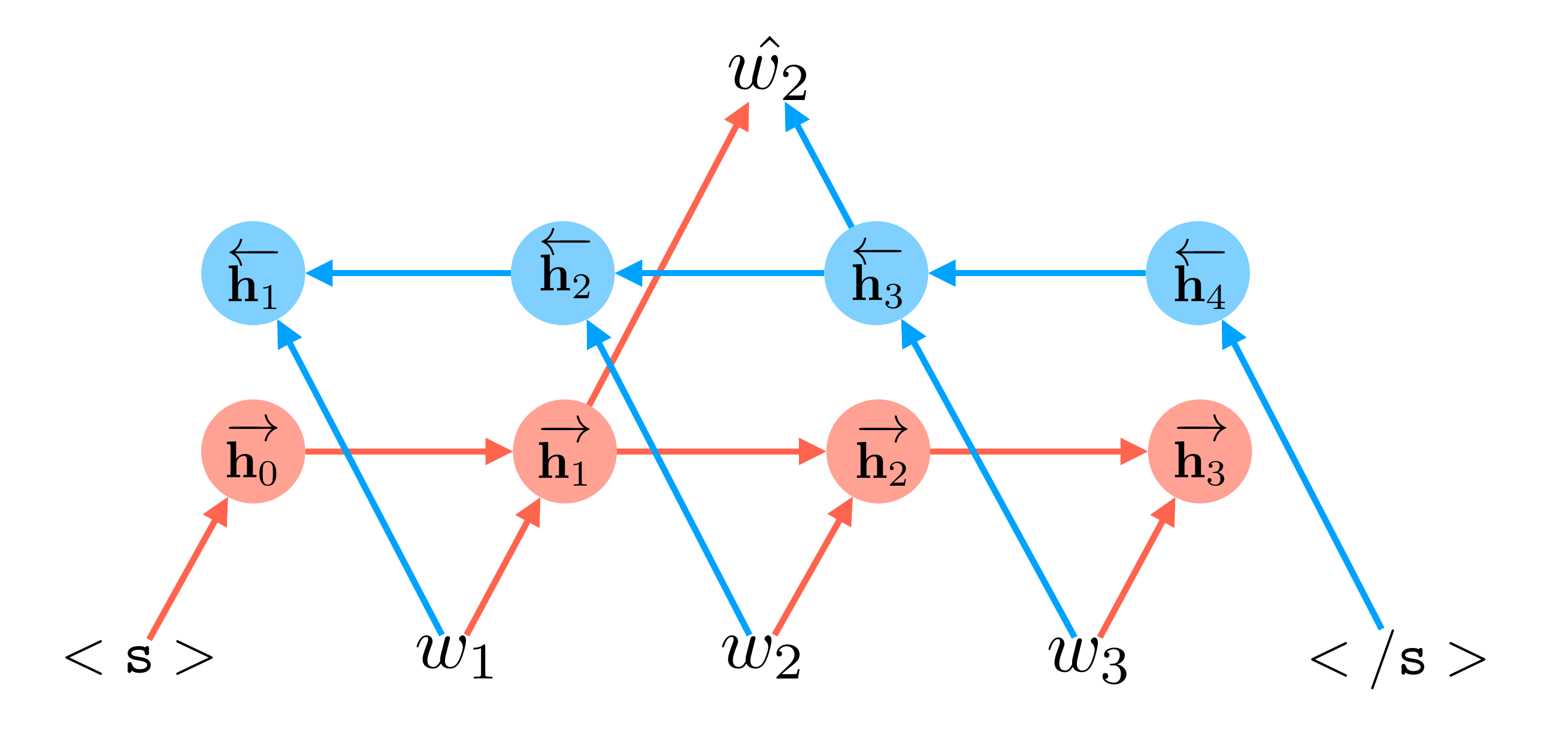}
}
\end{center}
\caption{Bidirectional Language Model}
\label{fig:BiLM}
\end{figure}

Fig.~\ref{fig:BiLM} shows the architecture of
the bidirectional LM \cite{mousa:2017}. The optimization goal of our LM
is to minimize the $i^{th}$ word's prediction error given the 
surrounding word's hidden state: 
\begin{equation}
p(w_i \mid \overrightarrow{\vech_{i-1}} \oplus \overleftarrow{\vech_{i+1}})
\label{eq:lm}
\end{equation}

\begin{figure}[t]
\begin{center}
\subfigure{
\includegraphics[height=5.5cm]{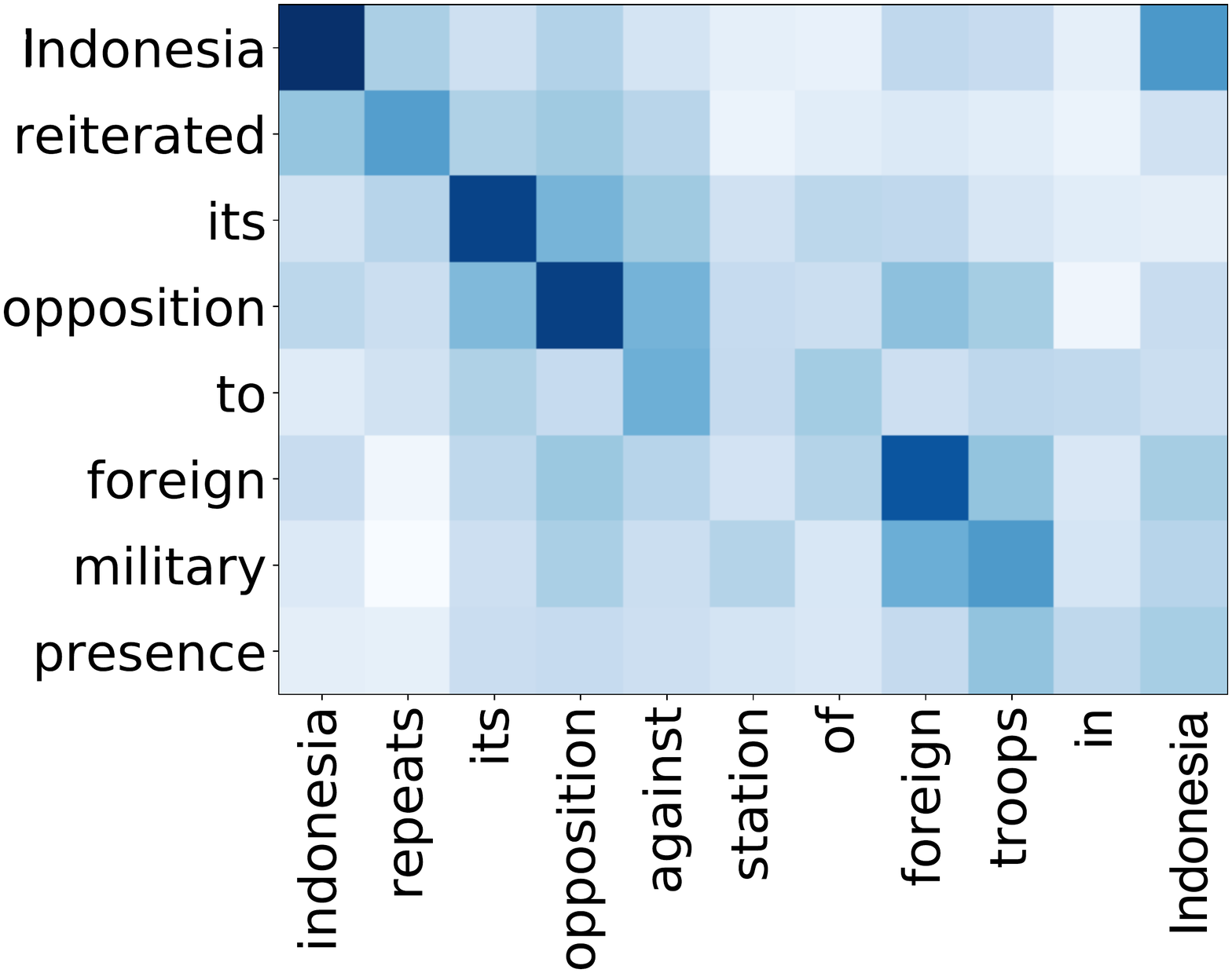}
}
\end{center}
\caption{Semantic Substitution Matrix}
\label{fig:matrix}
\end{figure}

For any new given sentences, we concatenate both forward and backward hidden states to represent each word $\vecy_{i, u}$ in a sentence $\vecy_i$.
We then calculate the normalized \textit{cosine} similarity score of word $\vecy_{i, u}$ and $\vecy_{j, v}$ as:

\vspace{-4ex}
\begin{equation}
M_{u, v} = cosine( \overrightarrow{\vech_u} \oplus \overleftarrow{\vech_u}, \overrightarrow{\vech_v} \oplus \overleftarrow{\vech_v})
\vspace{-1ex}
\end{equation}

Fig.~\ref{fig:matrix} shows an example of the semantic substitution matrix of first two sentences in example references of Fig.~\ref{fig:lattice_init}.


%% file: dp.tex

\subsection {Iterative Pairwise Word Alignment using Dynamic Programming}

With the help of semantic substitution matrix $M_{u,v}$
which measures pairwise word similarity,
we need to find the optimal word alignment to compress
references into a lattice.



Unfortunately, this computation is exponential in the number of sequences.
Thus, we use iterative pairwise alignment which greedily
merges sentence pairs \cite{Durbin:98}.

\begin{figure}[t]
\begin{center}
\subfigure{
\includegraphics[height=5.5cm]{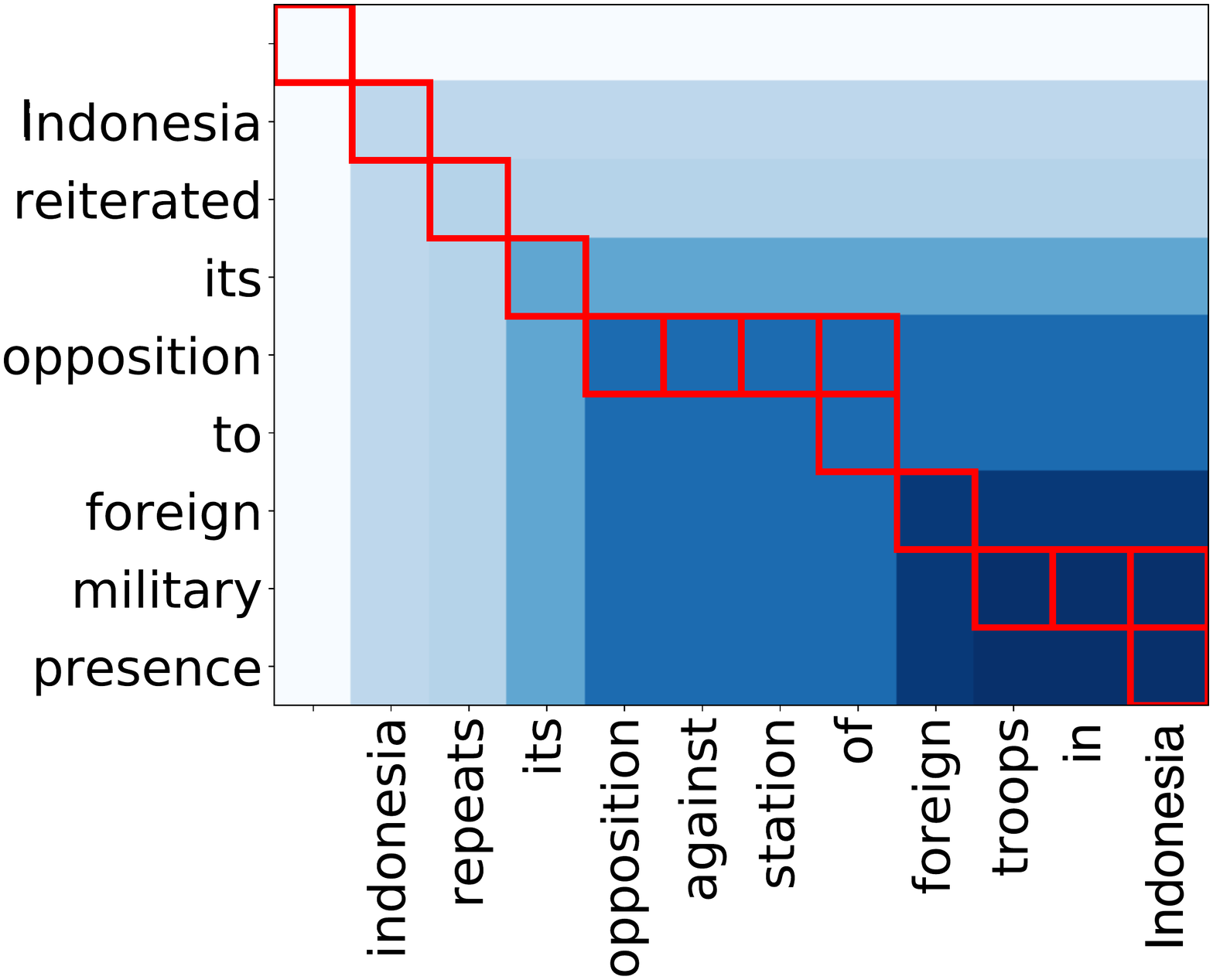}
}
\end{center}
\caption{Dynamic Programming on Semantic Substitution Matrix}
\label{fig:dp}
\end{figure}

Based on pairwise substitution matrix we can define an optimal pairwise sequence alignment as an optimal path from $M_{0, 0}$
to $M_{|\vecy_i|, |\vecy_j|}$.
This is a dynamic programming problem
with the state transition function described in Equation (\ref{eq:dp}).
Fig.~\ref{fig:dp} shows the optimal path according to
the semantic substitution matrix in Fig.~\ref{fig:matrix}.
There is a gap if the continuous step goes vertical or horizontal,
and an alignment if it goes diagonal.

\begin{equation}
opt(u, v)\!=\!
   \begin{cases}
   opt(u\!-\! 1, v\!-\! 1) \!+\! M_{u, v}\\
   opt(u\!-\! 1, v) \\
   opt(u, v\!-\! 1) 
   \end{cases}
\label{eq:dp}
\end{equation}

What order should we follow to do the iterative pairwise word alignment?
Intuitively, we need to compress the most similar reference pair first,
since this compression will lead to more aligned words.
Following this intuition, we order reference pairs by
the maximum alignment score $opt(|\vecy_i|, |\vecy_j|)$
(i.e. the score of bottom-right cell in Fig.~\ref{fig:dp})
which is the sum of all aligned words.
Using this order, we can iteratively merge
each sentence pair in descending order,
unless both the sentences have already been merged
(this will prevent generating a cyclic lattice).

Since the semantic substitution matrix $M_{u, v}$,
defined as a normalized cosine similarity, scales in $(0, 1)$,
it's very likely for the DP algorithm to align unrelated words.
To tackle this problem, we deduct a global penalty $p$ from 
each cell of $M_{u, v}$.
With the global penalty $p$,
the DP algorithm will not align a word pair $(\vecy_{i,u}, \vecy_{i,v})$
unless $M_{u, v} \geq p$.

After the pairwise references alignment,
we merge those aligned words.
For example, in Fig.~\ref{fig:lattice}, after we generate an initial
lattice as shown in Fig.~\ref{fig:lattice_init},
we then calculate the maximum alignment score of all sentence pairs.
After that, the lattice turns into Fig.~\ref{fig:latticeb}
by merging the first two references
(assuming they have the highest score) according to pairwise alignment
shown in Fig.~\ref{fig:dp}.
Then we pick the sentence pair with next highest alignment score
(assuming it's the last two sentences).
Similar to the previous step, we find alignments according to the 
dynamic programming and merge to the final lattice (see Fig.~\ref{fig:latticec}).

%% file: generate.tex

\subsection{Traverse Lattice and Pseudo-References Selection by BLEU}
\label{sec:generate}

We generate pseudo-references by simply traversing the generated lattice.
For example, if we traverse the final lattice shown in Fig.~\ref{fig:latticec},
we can generate 213 pseudo-refrences
in total.



Then, we can put those generated pseudo-references to
expand the training dataset.
To balance the number of generated pseudo-references for
each example, we force the total number of pseudo-references
from each example to be $K'$.
For those examples generating $k$ pseudo-references and $k > K'$,
we calculate all pseudo-references' BLEU scores based on gold references,
and only keep top $K' - k$ pseudo-references with highest BLEU score.

%% file: exps.tex

\section{Experiments}
\label{sec:experiments}

\begin{table*}[!htbp]
\centering
\resizebox{0.9\textwidth}{!}{
\begin{tabular}{|l|l|c|c|c|c|} \hline
Task                                 &                        & Pre-training & Training & Validation & Testing \\ \hline
\multirow{2}{*}{Machine Translation} & \# of examples         & 1,000,000    & 4,667    & 616        & 691   \\ \cline{2-6}
                                     & \# of refs per example & 1            & 4        & 4          & 4     \\ \hline
\multirow{2}{*}{Image Captioning}    & \# of examples         & -            & 113,287  & 5,000      & 5,000 \\ \cline{2-6}
                                     & \# of refs per example & -            & 5        & 5          & 5     \\ \hline
\end{tabular}}
\caption{Statistics of datasets used in following experiments.}
\label{tab:dataset}
\end{table*}

\begin{table}[t]
\centering
\resizebox{0.45\textwidth}{!}{
\begin{tabular}{|c|l|c|c|} \hline
\# of Refs                     & Method& \multicolumn{2}{c|}{BLEU}       \\ \hline
0                     & Pre-train              & \multicolumn{2}{c|}{37.44}          \\ \hline
1                     & $\text{First}^*$       & \multicolumn{2}{c|}{38.64}          \\ \hline
\multirow{3}{*}{4 }  & Sample One              & \multicolumn{2}{c|}{38.81}         \\
                               & Uniform       & \multicolumn{2}{c|}{38.78}  \\
                               & Shuffle       & \multicolumn{2}{c|}{38.87}           \\ \hline
\multicolumn{2}{|c|}{Includes Pseudo-Refs}        & Hard Align& Soft Align       \\ \hline
\multirow{3}{*}{10 }& Sample One & 37.48 & 39.41           \\
                               & Uniform       & 39.20 & 39.35          \\
                               & Shuffle       & 39.13 & 39.53          \\ \hline
\multirow{3}{*}{20 } & Sample One& 37.27 & 38.70          \\
                               & Uniform       & 39.14 & 39.46          \\
                               & Shuffle       & 39.12 & 39.42      \\ \hline
\multirow{3}{*}{50 } & Sample One& 37.42 & 37.62          \\
                               & Uniform       & 39.30 & \textbf{39.65}\\
                               & Shuffle       & 38.98     & 39.08          \\ \hline
\multirow{3}{*}{100 }&Sample One & 37.54 & 37.63          \\
                               & Uniform       & 39.23 & 39.46  \\
                               & Shuffle       & 38.88 & 39.03          \\ \hline
\end{tabular}}
\caption{BLEU on the MT validation set. {$^*$} Baseline}
\label{tab:mt_dev_result}
\end{table}

\begin{table}[t]
\centering\setlength{\tabcolsep}{3pt}
\begin{tabular}{|c|l|l|} \hline
\# of Refs    & Method& BLEU           \\ \hline
0 & Pre-train     & 33.58          \\ \hline
1 & $\text{First}^*$& 34.49          \\ \hline
4 & Shuffle       & 35.20 (+0.7)          \\ 
$\text{}^\dag$50 & Uniform       & \textbf{35.98} (+1.5)          \\ \hline
\end{tabular}
\caption{BLEU on the MT test set. {$^\dag$}Includes pseudo-references generated by soft word alignment algorithm. {$^*$}Baseline.}
\label{tab:mt_test_result}
\end{table}

\begin{table}[t]
\centering\setlength{\tabcolsep}{3pt}
\resizebox{0.48\textwidth}{!}{
\begin{tabular}{|c|l|cc|cc|} \hline
\# of Refs                     &       Method  & \multicolumn{2}{c|}{BLEU}  &  \multicolumn{2}{c|}{CIDEr}  \\ \hline
1                     & First                  & \multicolumn{2}{c|}{26.27} &  \multicolumn{2}{c|}{79.05}         \\ \hline
\multirow{3}{*}{5 }  & $\text{Sample One}^*$   & \multicolumn{2}{c|}{29.03} &  \multicolumn{2}{c|}{85.39}         \\
                               & Uniform       & \multicolumn{2}{c|}{30.05} &  \multicolumn{2}{c|}{89.76}    \\
                               & Shuffle       & \multicolumn{2}{c|}{30.41} &  \multicolumn{2}{c|}{91.21}    \\ \hline

 \multicolumn{2}{|c|}{\multirow{2}{*}{Includes Pseudo-Refs}}   & \multicolumn{2}{c|}{Hard Align}  &  \multicolumn{2}{c|}{Soft Align}  \\ 
 \multicolumn{2}{|c|}{ }                       & BLEU & CIDEr & BLEU     & CIDEr  \\ \hline
\multirow{3}{*}{10 } & Sample One& 30.63& 91.76 & 30.98    & 92.02         \\
                               & Uniform       & 30.40& 91.48 & 30.77    & 91.89      \\
                               & Shuffle       & 30.68& 92.01  & 30.91    & 92.22  \\ \hline
\multirow{3}{*}{20 } & Sample One& 30.69 & 92.25    & 30.91    & 92.32      \\
                               & Uniform       & 30.73& 91.69     & 31.03    & 92.61    \\
                               & Shuffle       &  31.56   & 94.99     & \textbf{31.92} & \textbf{95.59}       \\ \hline
\multirow{3}{*}{50 } & Sample One& 30.76    & 91.81     & 31.07    & 92.17   \\
                               & Uniform       & 30.66    & 92.30     & 30.99    & 92.61      \\
                               & Shuffle       & 30.83    & 93.26     & 31.06    & 94.19     \\ \hline
\end{tabular}
}
\caption{BLEU/CIDEr on the image captioning validation set. $^*$Baseline.}
\label{tab:caption_dev_result}
\end{table}

\begin{table}[t]
\centering\setlength{\tabcolsep}{3pt}
\resizebox{0.48\textwidth}{!}{
\begin{tabular}{|c|l|ll|} \hline
\# of Refs     & Method & BLEU      & CIDEr      \\ \hline
1 & First                  & 26.70     & 80.70      \\ \hline
5 & $\text{Sample One}^*$  & 28.67     & 85.41      \\ \hline
5 & Shuffle                & 30.94 (+2.3) & 94.10 (+8.7)     \\ 
$\text{}^\dag$20 & Shuffle & \textbf{31.79} (+3.1) & \textbf{97.10} (+11.7) \\ \hline
\end{tabular}
}
\caption{BLEU/CIDEr on the image captioning test set with soft. {$^\dag$} Includes pseudo-references generated by soft word alignment algorithm. {$^*$} Baseline.}
\label{tab:caption_test_result}
\end{table}

\begin{figure}[!t]
\centering
\begin{tabular}{c}
\begin{minipage}[t]{1.0 \linewidth}
\begin{center}
\subfigure[Machine Translation Dataset]{
\includegraphics[height=5.5cm]{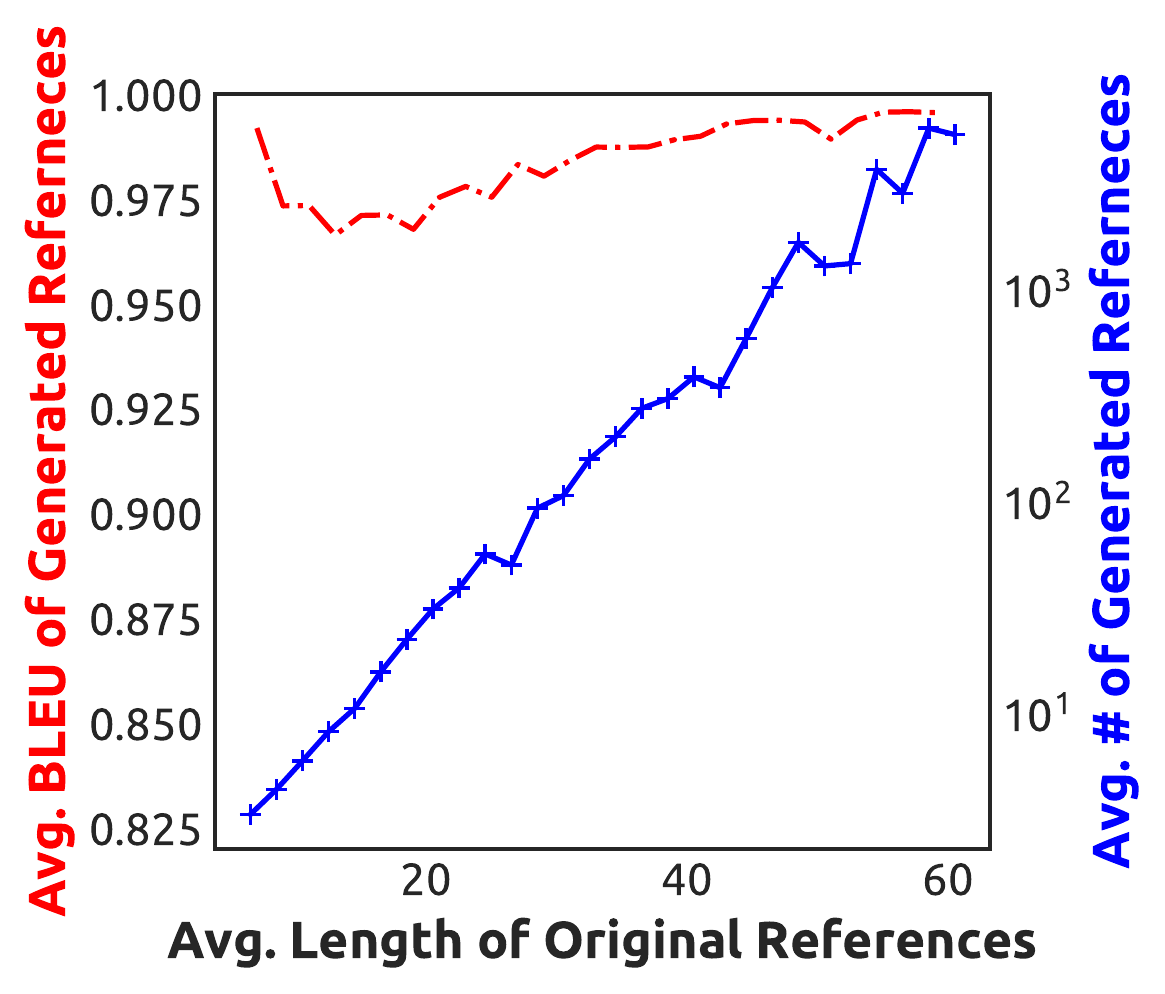}
\label{fig:ref-num-mt}
}
\end{center}
\end{minipage}
\\
\begin{minipage}[t]{1.0 \linewidth}
\begin{center}
\subfigure[Image Captioning Dataset]{
\includegraphics[height=5.5cm]{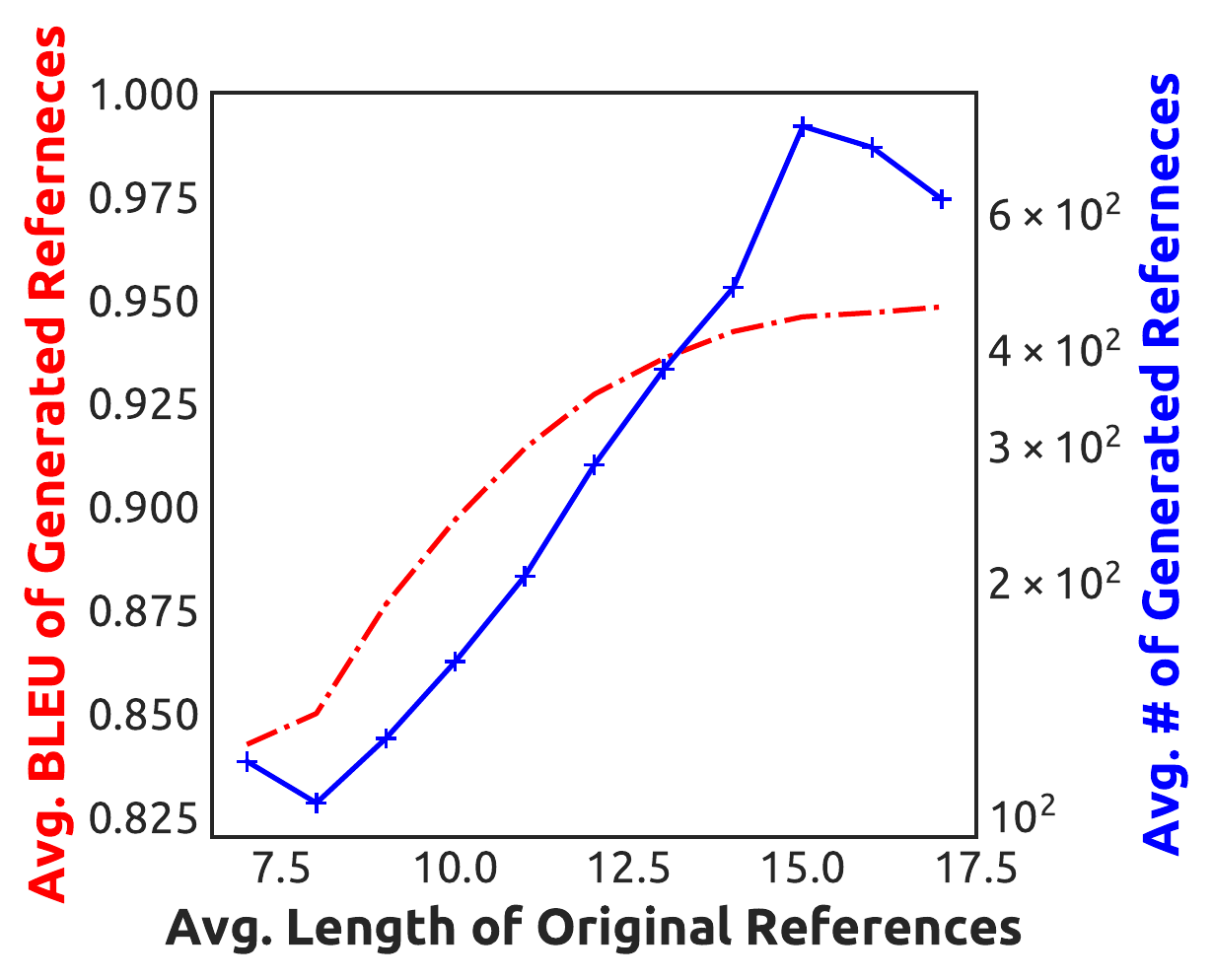}
\label{fig:ref-num-caption}
}
\end{center}
\end{minipage}
\end{tabular}
\caption{Analysis of generated references}
\end{figure}

To investigate the empirical performances of our proposed algorithm,
we conduct experiments on machine translation and image captioning.

\subsection{Machine Translation}

We evaluate our approach on NIST{} Chinese-to-English translation dataset
which consists of 1M pairs of single reference data and 5974
pairs of 4 reference data (NIST 2002, 2003, 2004, 2005, 2006, 2008).
Table \ref{tab:dataset} shows the statistics of this dataset.
We first pre-train our model on a 1M pairs single reference dataset
and then train on the NIST 2002, 2003, 2004, 2005.
We use the NIST 2006 dataset as validation set and NIST 2008 as test sets.

Fig.~\ref{fig:ref-num-mt} analyzes the number and quality of
generated references using our proposed approach.
We set the global penalty as 0.9 and only calculate
the top 50 generated references for the average BLEU analysis.
From the figure, we can see that when the sentence length grows,
the number of generated references grows exponentially.
To generate enough references for the following experiments,
we set an initial global penalty as 0.9 and gradually decrease it by 0.05
until we collect no less than 100 references.
We train a bidirectional language model on the
pre-training dataset and training dataset
with Glove \cite{pennington2014glove} word embedding size of 300 dimension,
for 20 epochs to minimize the perplexity

We employ byte-pair encoding (BPE) \cite{sennrich2015neural} which
reduces the source and target language vocabulary sizes to 18k and 10k.
We adopt length reward \cite{huang+:2017} to find
optimal sentence length.
We use a two layer bidirectional LSTM as the encoder and a two
layer LSTM as the decoder.
We perform pre-training for 20 epochs to minimize perplexity on the
1M dataset, with a batch size of 64, word embedding size of 500,
beam size of 15, learning rate of 0.1, learning rate decay of 0.5
 and dropout rate of 0.3.
We then train the model in 30 epochs and use the best batch size among
100, 200, 400 for each update method.
These batch sizes are multiple of the number of references
used in experiments, so it is guaranteed that all the references of
one single example are in one batch for the Uniform method.
The learning rate is set as 0.01 and learning rate decay as 0.75.
We do each experiment three times and report the average result.

\begin{figure}[!hbt]
\centering
\begin{tabular}{c}
\begin{minipage}[t]{1.0 \linewidth}
\begin{center}
\subfigure[Learning curve of different methods with 50 References]{
\includegraphics[height=5.0cm]{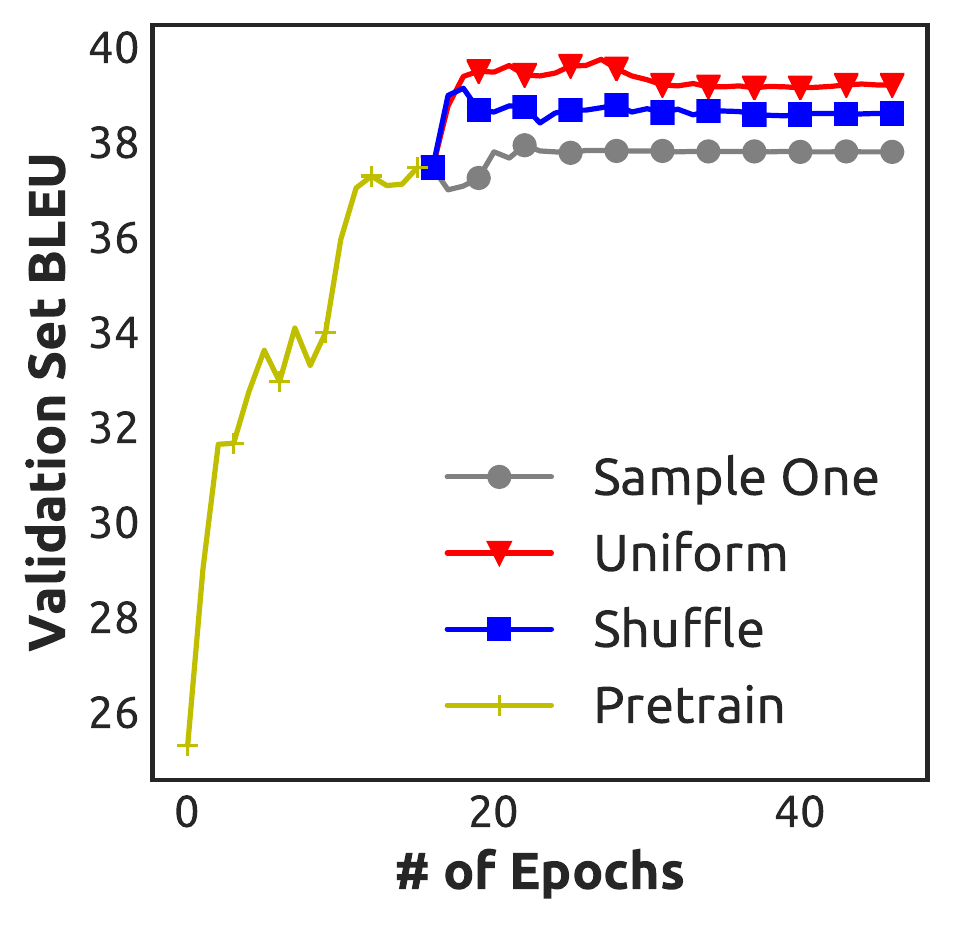}
\label{fig:mt-curve-methods}
}
\end{center}
\end{minipage}
\\
\begin{minipage}[t]{1.0 \linewidth}
\begin{center}
\subfigure[MT with different number of references]{
\includegraphics[height=5.0cm]{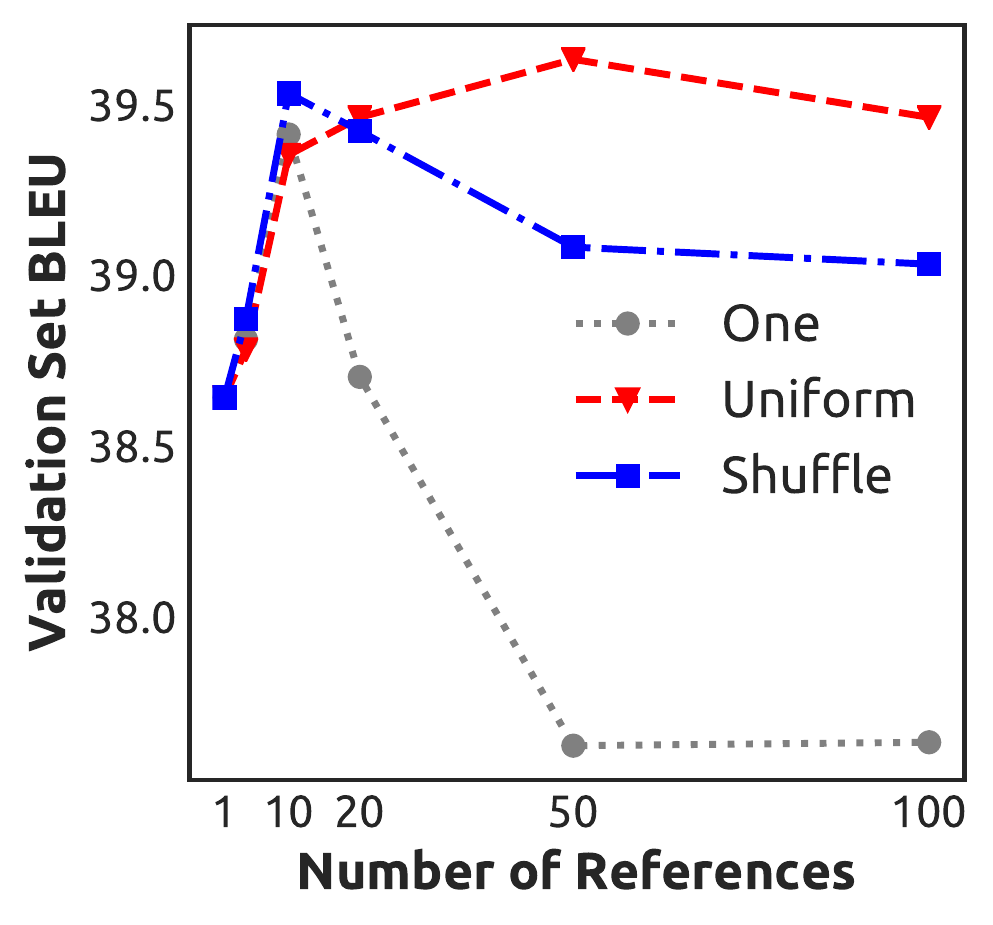}
\label{fig:mt-curve-bleu}
}
\end{center}
\end{minipage}
\vspace{-15pt}
\end{tabular}
\caption{Translation quality of machine translation task on dev-set with soft alignment}
\vspace{-15pt}
\end{figure}

Table \ref{tab:mt_dev_result} shows the translation quality
on the dev-set of machine translation task.
Besides the original 4 references in the training set,
we generate another four dataset with 10, 20,
 50 and 100 references including pseudo-references
 using hard word alignment and soft word alignment.
We compare the three update methods
(Sample One, Uniform, Shuffle)
with always using the first reference (First).
All results of soft word alignment are better than corresponding hard word alignment results
and the best result is achieved with 50 references using Uniform and soft word alignment.
According to Table \ref{tab:mt_test_result}, Shuffle with original 4
references has +0.7 BLEU improvement and Uniform with 50 references has
+1.5 BLEU improvement.
From Fig.~\ref{fig:mt-curve-bleu}, we can see that using the
Sample One method, the translation
quality drops dramatically with more than 10 references.
This may be due to the higher variance of used reference in
each epoch.


\begin{figure}[!htb]
\centering
\begin{tabular}{c}

\begin{minipage}[t]{1.0 \linewidth}
\begin{center}
\subfigure[Learning curve of different methods with 20 References]{
\includegraphics[height=5.5cm]{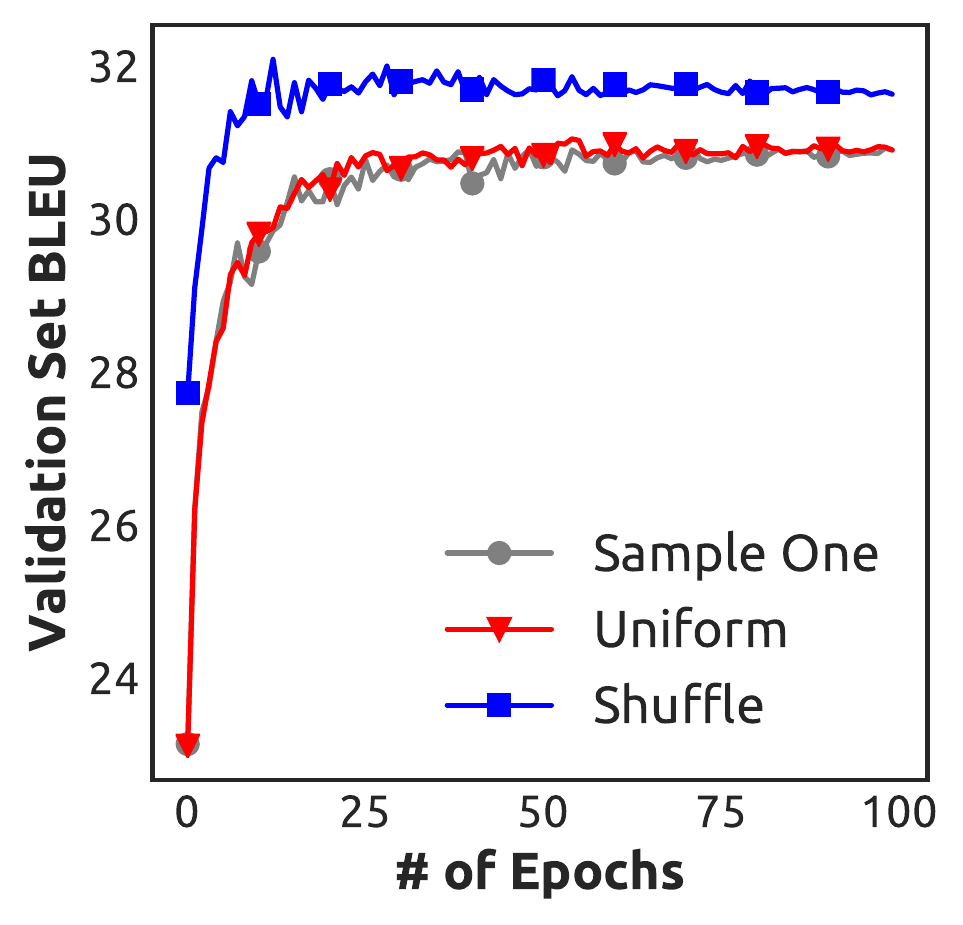}
\label{fig:caption-curve-methods}
}
\end{center}
\end{minipage}
\\
\begin{minipage}[t]{1.0 \linewidth}
\begin{center}
\subfigure[Image captioning with different number of references]{
\includegraphics[height=5.5cm]{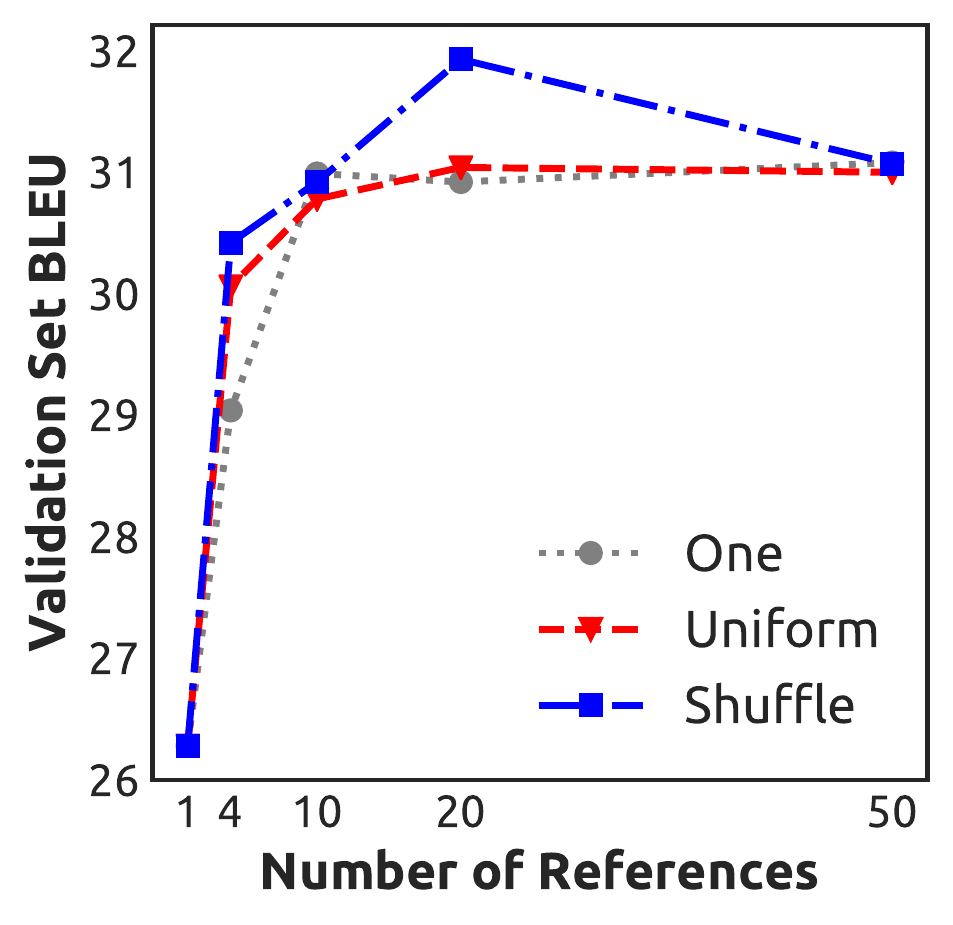}
\label{fig:caption-curve-bleu}
}
\end{center}
\end{minipage}
\end{tabular}
\caption{Text generation quality of image captioning task
on validation set with soft alignment}
\end{figure}

\begin{table*}[!htb]

\centering
\resizebox{1.0\textwidth}{!}{
\begin{tabular}{|l|p{15.5cm}|}
 \hline
Image          & Original References \\ \hline

\begin{minipage}{.25\textwidth}
\vspace{0.3cm}
    \includegraphics[width=\linewidth]{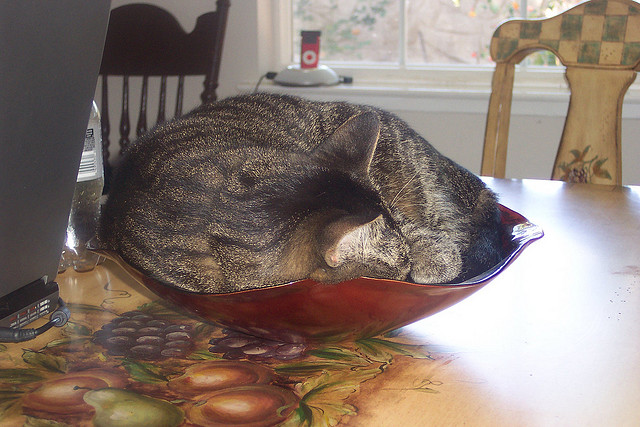}
\end{minipage}
& 
\vspace{-1.9cm}
{\small
\hspace{-1.7cm}
\begin{enumerate}
\itemsep0em 
\item[] \texttt{a gray tabby cat is curled in a red bowl that sits on a table near a window}
\item[] \texttt{a brown and black cat is sleeping in a bowl on a table}
\item[] \texttt{a grey tiger cat sleeping in a brown bowl on a table}
\item[] \texttt{an image of a cat sitting inside of a bowl on the kitchen table}
\item[] \texttt{a cat asleep in a fruit bowl on a dining room table}
\end{enumerate}
}
\\ \hline

\multicolumn{2}{|c|}{Generated Lattice using Soft Alignment}

\\ \hline

\multicolumn{2}{|c|}{
\begin{minipage}{1.2\textwidth}
    \includegraphics[width=\textwidth]{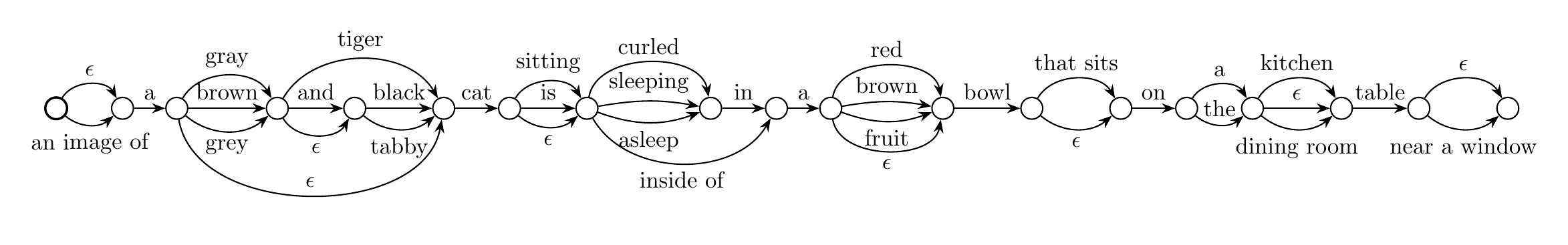}
\end{minipage}
}
\\ \hline
\multicolumn{2}{|c|}{
\begin{tabular}{r|l|r}
ID & Pseudo-references & BLEU \\ 
1 & $\texttt{a grey tiger cat sleeping in a brown bowl on a table near a window}$ & 100.0  \\
2 & $\texttt{a grey tiger cat sleeping in a brown bowl on a dining room table}$ & 100.0  \\
3 & $\texttt{a brown and black cat is sleeping in a bowl on the kitchen table}$ & 100.0  \\
... & $\texttt{...}$ & ...  \\
48 & $\texttt{a grey tiger cat sleeping in a fruit bowl on a table}$ & 97.1  \\
49 & $\texttt{a cat asleep in a red bowl that sits on a table}$ & 97.1  \\
50 & $\texttt{a gray tabby cat is sleeping in a bowl on a table}$ & 97.1  \\
... & $\texttt{...}$ & ...  \\
73723 & $\texttt{a grey and tabby cat inside of a red bowl on the dining room table}$ & 0.0  \\
73724 & $\texttt{a grey and tabby cat inside of a red bowl on a kitchen table}$ & 0.0 \\
\end{tabular}
}
\\ \hline
\end{tabular}
}
\caption{Training example that generates maximum number of pseudo-references (73724). The selected 8 pseudo-references are sorted according to their BLEU score.}
\label{tab:caption_examples}
\vspace{-15pt}
\end{table*}

\subsection{Image Captioning}

For the image captioning task, we use the widely-used MSCOCO image captioning dataset.
Following prior work, we use the Kapathy split \cite{karpathy2015deep}.
Table \ref{tab:dataset} shows the statistics of this dataset.
We use Resnet \cite{kaiming:2016} to extract image feature
of 2048 feature size and
simple fully connected layer of size 512 to an LSTM decoder.
We train every model for 100 epochs and calculate the BLEU
score on validation set and select the best model.
For every update method, we find the optimal batch size among 50, 250, 500,
1000 and we use a beam size of 5.

Fig.~\ref{fig:ref-num-caption} analyzes the correlation between
average references length with the number and quality of generated references.
We set global penalty as 0.6 (which is also adopted for
the generated references in the following experiments) and calculate
the top 50 generated references for the average BLEU analysis.
Since the length of original references is much shorter than the previous
machine translation dataset, it has worse quality and fewer generated references.

Table~\ref{tab:caption_dev_result} shows that the best result is 
achieved with 20 references using Shuffle.
This result is different from the result of machine translation task
where Uniform method is the best.
This may be because the references in image captioning dataset are much more
diverse than those in machine translation dataset.
Different captions of one image could even talk about different aspects.
When using the Uniform method, the high variance of references
in one batch may harm the model and lead to worse text generation quality.
Table~\ref{tab:caption_test_result} shows that it outperforms Sample One
with 4 original references, which is adopted in previous work \cite{karpathy2015deep}, +3.1 BLEU score and +11.7 CIDEr.

\subsection{Case Study}

Fig.~\ref{tab:caption_examples} shows a training example in 
the COCO dataset and its corresponding generated lattice and
pseudo-references which is sorted according to its BLEU score.
Our proposed algorithm generates 73724 pseudo-references in total.
All the top 50 pseudo-references' BLEU scores are above 97.1
and the top three even achieve 100.0 BLEU score though they are not
identical to any original references.
Although the BLEU of last two sentences is 0.0, they are still
valid to describe this picture.